\documentclass[10pt, twocolumn, letterpaper]{article}

\usepackage[pagenumbers]{cvpr} 

\usepackage[dvipsnames]{xcolor}

\definecolor{cvprblue}{rgb}{0.21,0.49,0.74}
\usepackage[pagebackref,breaklinks,colorlinks,citecolor=cvprblue]{hyperref}

\def\paperID{14227} 
\def\confName{CVPR}
\def\confYear{2024}

\title{\textbf{DGNet}: Dynamic Gradient-Guided Network for Water-Related Optics Image Enhancement}

\author{Jingchun Zhou\textsuperscript{\rm 1},
Zongxin He\textsuperscript{\rm 1}, 
Qiuping Jiang\textsuperscript{\rm 1}, 
Kui Jiang\textsuperscript{\rm 2}, 
Xianping Fu\textsuperscript{\rm 3}, \\
Xuelong Li\textsuperscript{\rm 4}, 
\\
\textsuperscript{\rm 1} Dalian Maritime Unverisity, 
\textsuperscript{\rm 2} Ningbo University,
\textsuperscript{\rm 3} Harbin Institute of Technology, \\
\textsuperscript{\rm 4} China Telecom Corp. Ltd.
}

\begin{document}
\maketitle
\begin{abstract}

Underwater image enhancement (UIE) is a challenging task due to the complex degradation caused by underwater environments. To solve this issue, previous methods often idealize the degradation process, and neglect the impact of medium noise and object motion on the distribution of image features, limiting the generalization and adaptability of the model. Previous methods use the reference gradient that is constructed from original images and synthetic ground-truth images. This may cause the network performance to be influenced by some low-quality training data.
Our approach utilizes predicted images to dynamically update pseudo-labels, adding a dynamic gradient to optimize the network's gradient space. This process improves image quality and avoids local optima. Moreover, we propose a Feature Restoration and Reconstruction module (FRR) based on a Channel Combination Inference (CCI) strategy and a Frequency Domain Smoothing module (FRS). These modules decouple other degradation features while reducing the impact of various types of noise on network performance. Experiments on multiple public datasets demonstrate the superiority of our method over existing state-of-the-art approaches, especially in achieving performance milestones: PSNR of 25.6dB and SSIM of 0.93 on the UIEB dataset. Its efficiency in terms of parameter size and inference time further attests to its broad practicality. The code will be made publicly available.

\end{abstract}    
\section{Introduction}
\label{sec:intro}

\begin{figure}[!t]
\centering 
\includegraphics[width=0.45\textwidth]{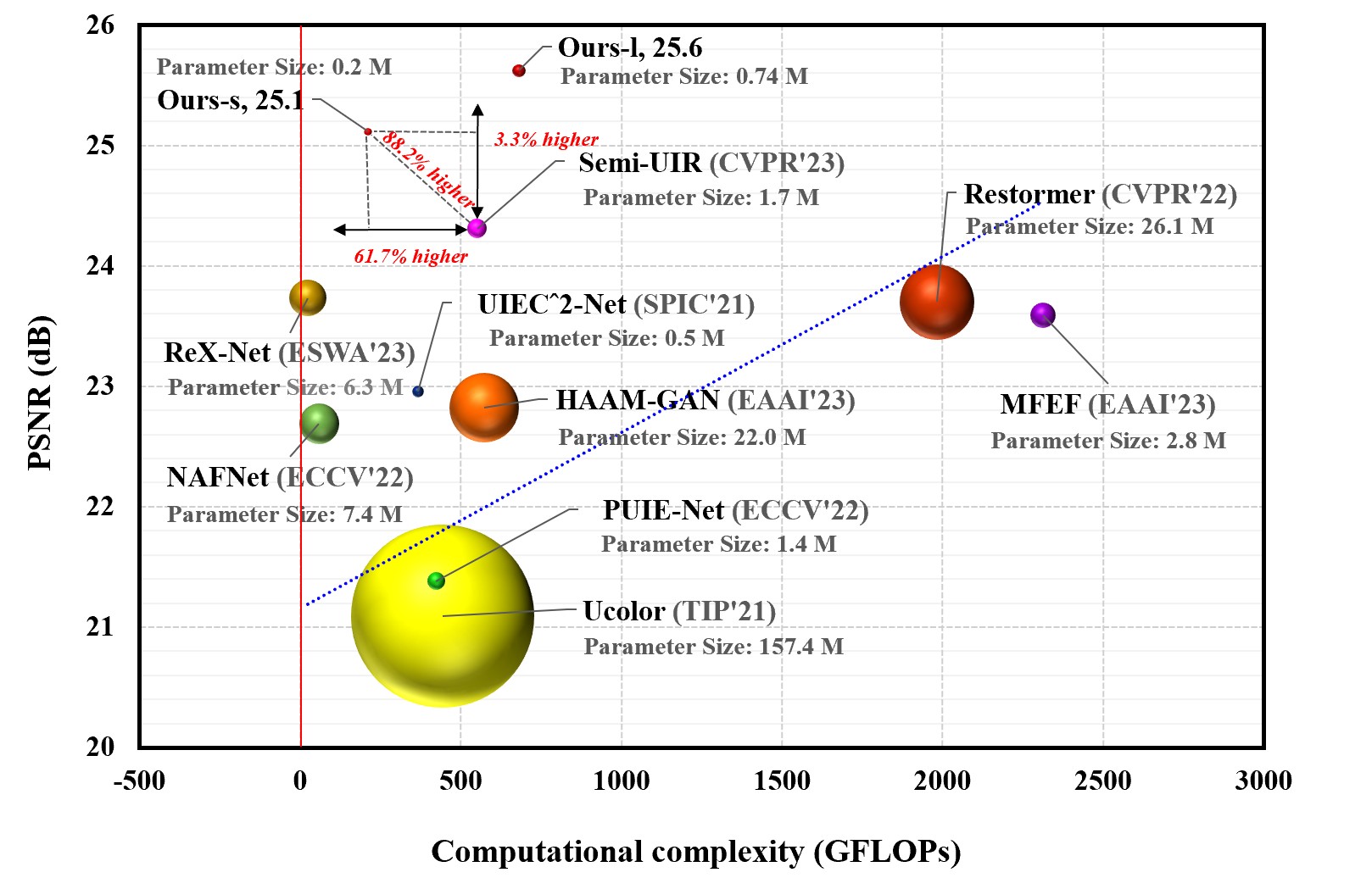}
\caption{Performance comparison between state-of-the-art methods \cite{R38ReX-Net, R36NAFNet, R26UIEC, R33PUIENet, R25Ucolor, R37SemiUIR, R39HAAMGAN, R35Restormer, R34MFEM} and our methods on the UIEB V90 dataset \cite{R18WaterNet}. Evaluation metrics include PSNR for image quality, computational complexity (GFLOPs), and parameter size represented by ball size. Our method achieves remarkable performance milestones. 
}
\label{Fig1}
\end{figure}

In recent years, deep learning-based methods have demonstrated their efficient computation and outstanding performance, leading to their widespread applications in computer vision, such as image denoising~\cite{R5}, deblurring~\cite{R6,R8,R10,R47LEDNet}, dehazing~\cite{R9,R12,R48Dehamer}, deraining~\cite{R13,R14,R15}, desnowing~\cite{R16,R17desnowing}, and enhancing images under low-light conditions~\cite{R7,R11,R49zerodce, R50li2021lliesurvey}. Deep learning makes significant breakthroughs. The advancements achieved in these areas markedly elevate the visual quality of images captured under adverse conditions, significantly bolstering the efficacy of downstream applications, such as object detection, classification, and image matching. However, most existing methods focus on specific types of image degradation, while underwater image enhancement (UIE) involves multiple degradation factors, such as noise, marine snow, and low-light conditions. This makes UIE more challenging and complex than single-degradation image enhancement. \par

Due to the involvement of multiple degradation factors, underwater degradation scenes are complex. The water medium absorbs and scatters, causing underwater images to have high noise levels, uneven illumination, low contrast, blurred details, and color shifts, among other visual problems~\cite{R1,R2}. These issues make it challenging to obtain clear optical information for underwater exploration tasks, which is a challenge for both human vision and computer vision. Most of the current deep learning-based UIE methods use supervised learning frameworks. These methods require paired datasets to learn the mapping from degraded images to clear images. However, due to the limitations of optical devices and the complexity of underwater imaging, obtaining clear underwater images or ideal reference images for restoration is a challenging task. 
Existing methods based on synthetic datasets suffer from the following issues. \textbf{(1) Limited paired real data:} artificially synthesized degraded images may not capture the full diversity and complexity of real underwater scenes \cite{R40WaterGAN}. These synthetic images cannot accurately simulate the real underwater conditions. Therefore, models trained using synthetic data cannot handle unseen or challenging underwater scenarios well. \textbf{(2) Unpredictable misdirection:} using the best results from existing enhancement methods as ground truth \cite{R18WaterNet}. 
This approach limits supervised learning methods to the performance of existing enhancement methods. It may also introduce unwanted semantic features and distort the original characteristics of the images \cite{R53rethinking}, resulting in models that cannot generalize well to real-world underwater images. These issues show the difficulties in creating datasets for UIE and the potential drawbacks of relying only on existing data for supervision.



In order to mitigate the impact of low-quality data in the dataset and address the complex degradation factors present in different underwater scenarios, we propose a novel method called DGNet, which stands for Dynamic Gradient-guided Network for UIE. The core idea of DGNet involves the use of self-updating pseudo-labels to construct additional dynamic gradients, coupled with the supervision of reference gradients in the backward propagation process. During the training process, these pseudo-labels are generated using the Contrast Limited Adaptive Histogram Equalization (CLAHE) method, imparting sensitivity to illumination and contrast changes. This sensitivity guides the network to focus more deeply on the fundamental causes of image degradation, extending beyond the constraints imposed by reference ground truth images. Moreover, the dynamic gradient adjusts the gradient space after each iteration, facilitating the network to escape the constraints of saddle points in the gradient space, which allows it to explore a broader solution space.

Furthermore, when confronted with the challenge of image enhancement under the influence of noise, we design two innovative modules to collectively mitigate the impact of various types of noise. Firstly, the Feature Restoration and Reconstruction module (FRR) eliminates irregular noise and other degradation factors by reconstructing the overall distribution of features, effectively addressing the majority of degradation issues. Secondly, the Frequency Domain Smoothing module (FRS) smooths motion noise and low-frequency features. By initially extracting high-frequency features, it guides attention to the low-frequency region, smoothing color variations in that region and alleviating issues such as artifacts.
The experimental results on the UIEB dataset, a public benchmark, show that our methods, Our-S (0.2M parameters, suitable for severely arithmetic-limited scenarios) and Our-L (0.7M parameters) achieve PSNR scores of 25.1dB and 25.6dB, respectively. These scores largely surpass the current state-of-the-art methods. Additionally, our methods have better parameter size and computational efficiency than the state-of-the-art methods (see Fig. \ref{Fig1} and Tab. \ref{T1}). This suggests their high generalization potential and broad applicability.\par
The contributions of this paper are summarized below. \par
(1) We present DGNet, a Dynamic Gradient-guided Network for UIE. Our network incorporates a pseudo-label strategy to construct additional dynamic gradients, guiding the network to delve deeper into the fundamental causes of image degradation. This enables DGNet to adapt to diverse scenarios. Furthermore, the pseudo-labels dynamically adjust the gradient space after each epoch, providing the network with increased momentum during a training cycle, and facilitating its exploration of a broader gradient space to seek optimal solutions.\par
(2) We design multiple modules to address the impact of noise. Unlike other denoising methods that are specifically tailored to address certain types of noise, our modules are not designed to tackle specific noise types. Instead, they focus on eliminating all degradation factors that deviate from the overall distribution. Consequently, these modules do not introduce additional parameters or computational load, ensuring the network's broad applicability and lower risk of overfitting. \par
(3) Our network features a simple and efficient architecture with a low parameter size, designed to address the practical challenges of underwater environments. Our small model operates at a processing speed of 26.3 frames per second for 480P images on an RTX 3090 GPU, achieving high efficiency without compromising performance. 

\section{Related Work}


\begin{figure}[tbp]
\centering 
\includegraphics[width=0.9\linewidth]{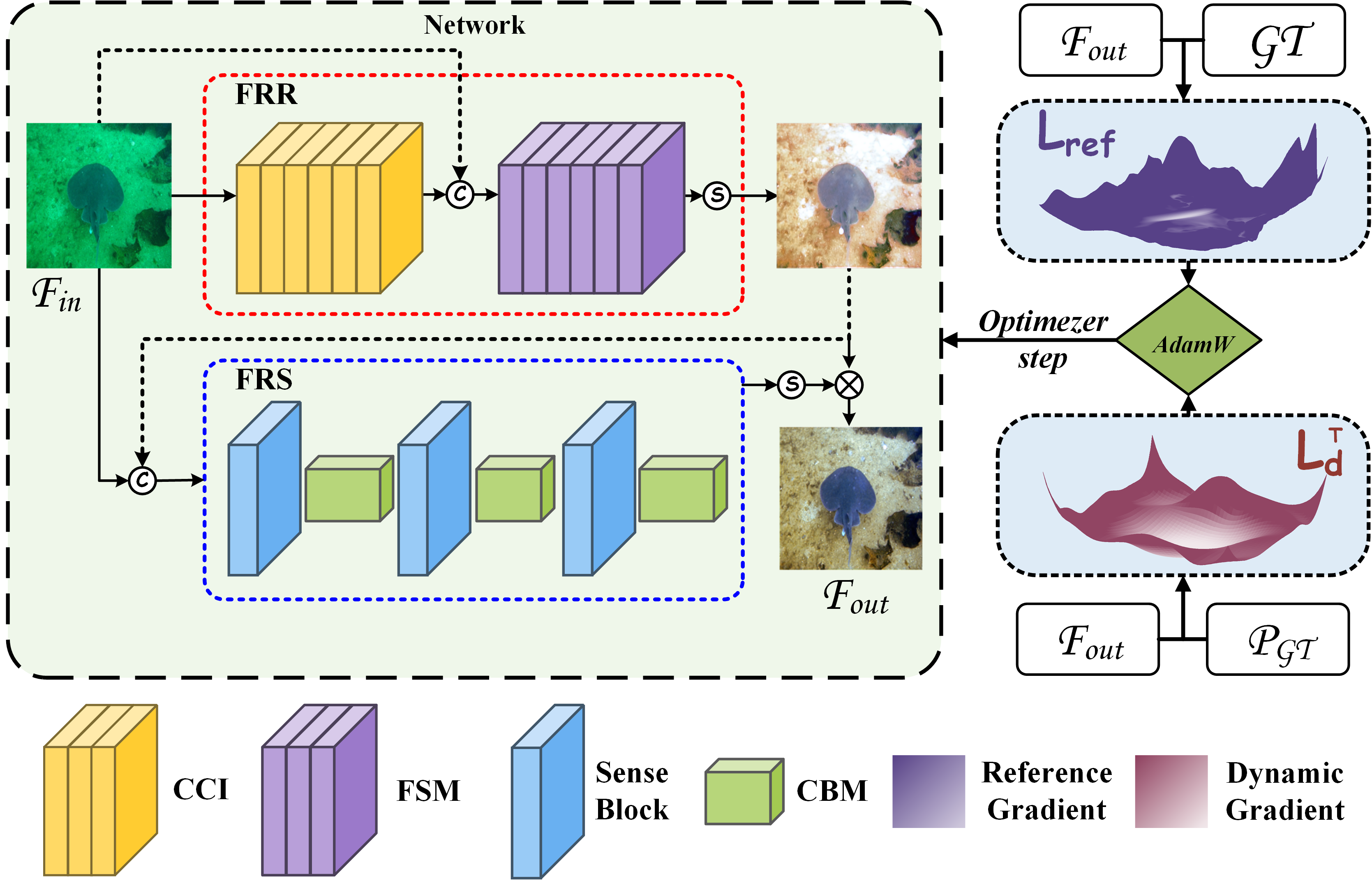}
\caption{Overall structure of DGNet. The FRR deals with noise and other degradation factors, while the FRS module smooths features. FRR consists of Channel Convolution Inference (CCI) and Feature Smoothing Module (FSM) (detail in Fig. \ref{FRR_detail}). We construct a dynamic gradient with the predicted image, which along with the reference gradient, jointly supervises the network. The CBM consists of a convolutional, BN layer with MiSH activation function. }
\label{Fig2}
\end{figure}

\subsection{Underwater Data-Driven Strategies}
The key to underwater vision tasks lies in how to acquire authentic and diverse semantic information, which is heavily dependent on the quality of existing underwater datasets. Existing underwater datasets can be categorized into two types: artificially synthesized \cite{R28UWCNN} and authentically collected \cite{R17desnowing}. The artificially synthesized datasets mostly consist of manually generated degraded terrestrial scenes, offering advantages such as abundant quantity and simple label creation. In contrast, authentically collected data can reflect the real rules of underwater imaging, but their quality is affected by the manual labeling process, and they are costly to acquire.

Based on the utilization of datasets, deep learning-based image enhancement methods can be classified into unsupervised, semi-supervised, and fully supervised approaches. Unsupervised methods like WaterGAN \cite{R40WaterGAN} and HybrUR \cite{R39HAAMGAN} employ Generative Adversarial Networks (GANs) to simulate realistic underwater scenes, alleviating the high cost of dataset creation. The semi-supervised Semi-UIR \cite{R4} method integrates supervised and contrastive learning, utilizing both labeled and unlabeled data for image enhancement. Supervised methods \cite{R25Ucolor,R51URanker,R41zhangdan} are limited by the distribution and quality of datasets, leading to suboptimal performance in specific scenarios.

The effectiveness of UIE methods is constrained by the quality of datasets, making it challenging to address this issue through data acquisition methods alone. Unsupervised approaches have trouble verifying the accuracy of the learned imaging relationships, while supervised approaches suffer from fitting errors due to low-quality real data.

\subsection{Lightweight and Effective UIE Method}
UIE is a crucial task for underwater exploration robots, which have limited computational resources that they also need for other functions. Traditional methods \cite{R41zhangdan, R42song, R31MLLE, R30GDCP, R52Liwater16} achieve favorable results in specific scenarios. They usually separate the degradation process into steps based on color bias, noise, and brightness, but do not consider how enhancements affect computer vision perception. Deep learning-based methods generally use underwater imaging models to describe the imaging mechanism, focusing on subjective factors like color bias and brightness, while neglecting noise issues caused by backscattering. Some methods use data augmentation to deal with noise, but most of them do not match the real noise conditions well.

In summary, many approaches fail to improve complex situations that involve several interrelated deterioration factors. They focus on prominent factors like brightness and color bias, neglecting the impact of noise on machine vision. Some methods with high parameter sizes \cite{R45UShape} or low efficiency are ill-suited for underwater conditions.

\section{Proposed Method}
In this section, we describe our network and components, where the FRR module adept at addressing complex degradation factors (e.g., irregular noise, illumination, color bias, etc.), while the FRS module is used to deal with remove motion noise occurring underwater to avoid artifacts, and to accelerate the training of the network to cope with the low-quality images in the dataset by using dynamic gradients constructed from pseudo-labels to direct the network's attention to the deep degradation features.

\begin{figure}[!t]
\centering 
\includegraphics[width=3in]{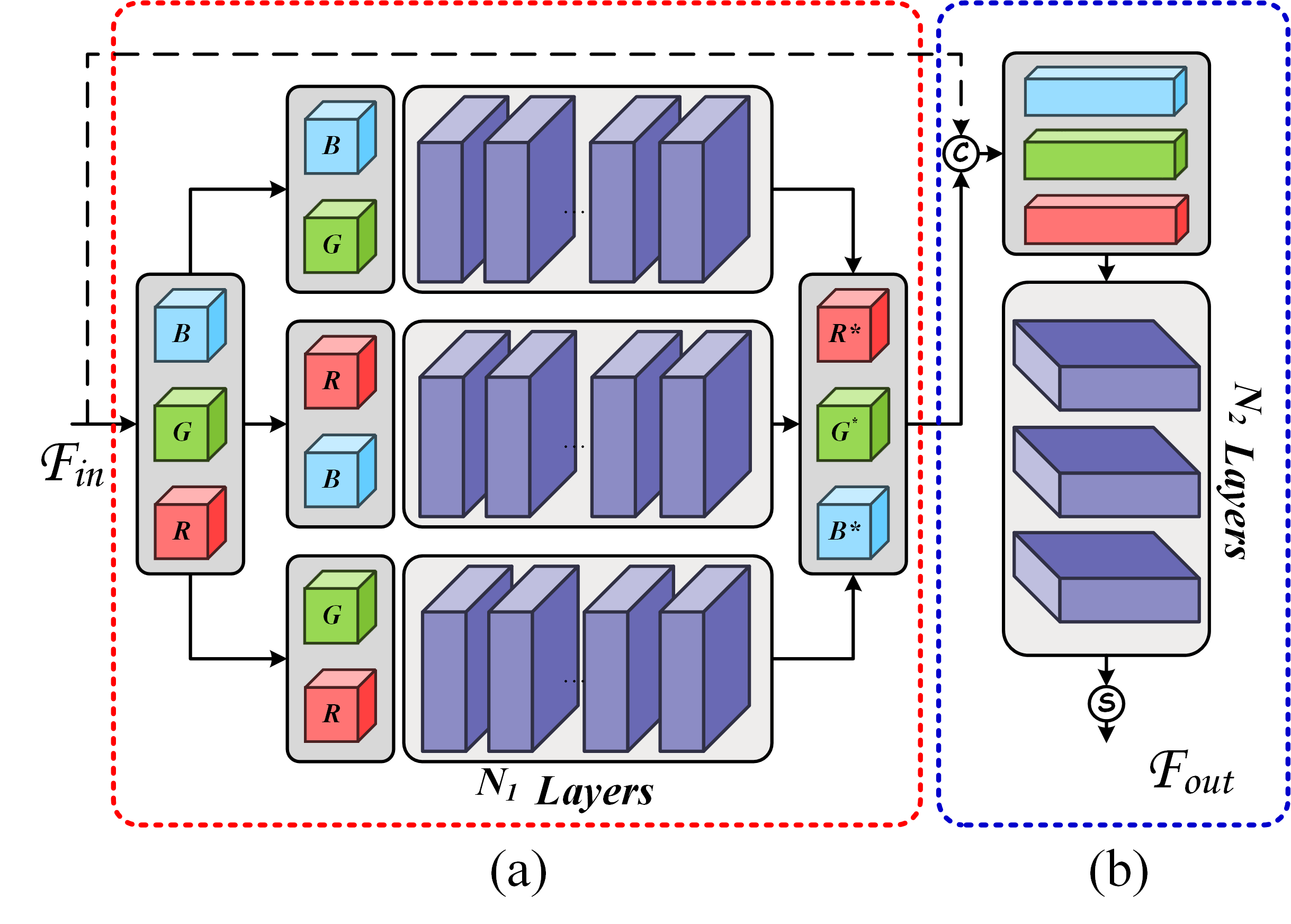}
\caption{Detail in FRR module. (a) Channel Combination Inference (CCI). (b) Fusion Sense Module (FSM). The \textcolor[RGB]{105, 105, 180}{blue block} is the CBM that consists of a convolution, BN layer with MiSH activation function. In the CCI block, we replace normal convolution with group convolution.}
\label{FRR_detail}
\end{figure}

\subsection{Feature Reconstruction and Re-Learning}
Underwater degradation factors are complex and varied, making it challenging for networks to adapt their diverse characteristics. Most existing underwater methods tend to ignore noise \cite{R32SMBL, R46PUGAN}, or use more computational resources to learn the noise distribution \cite{R43TGRS}. In practical training, the networks with large amounts of parameters can lead to overfitting. We think that this inefficient strategy fails to eliminate erroneous representations and does not serve downstream high-level vision tasks effectively.

To address the above issues, we propose an innovative channel pixel inference strategy, analyzing the underwater imaging model and the image sampling mechanism. In the RGB image, each channel represents the same scene with different statistical characters. When there is no noise interference, the feature expression conforms to the light attenuation model; otherwise, undesired features are introduced into the RGB channels. Our designed channel pixel inference strategy takes advantage of this characteristic, that is, any pixel point other than the primary colors should simultaneously exist in two channels. When we remove a certain channel pixel, the network infers the true pixel through the learned attenuation model. Even for primary color pixels, which only exist in one channel, the network can simply infer them from the adjacent local area. 

\begin{figure}[!t]
\centering 
\includegraphics[width=0.4\textwidth]{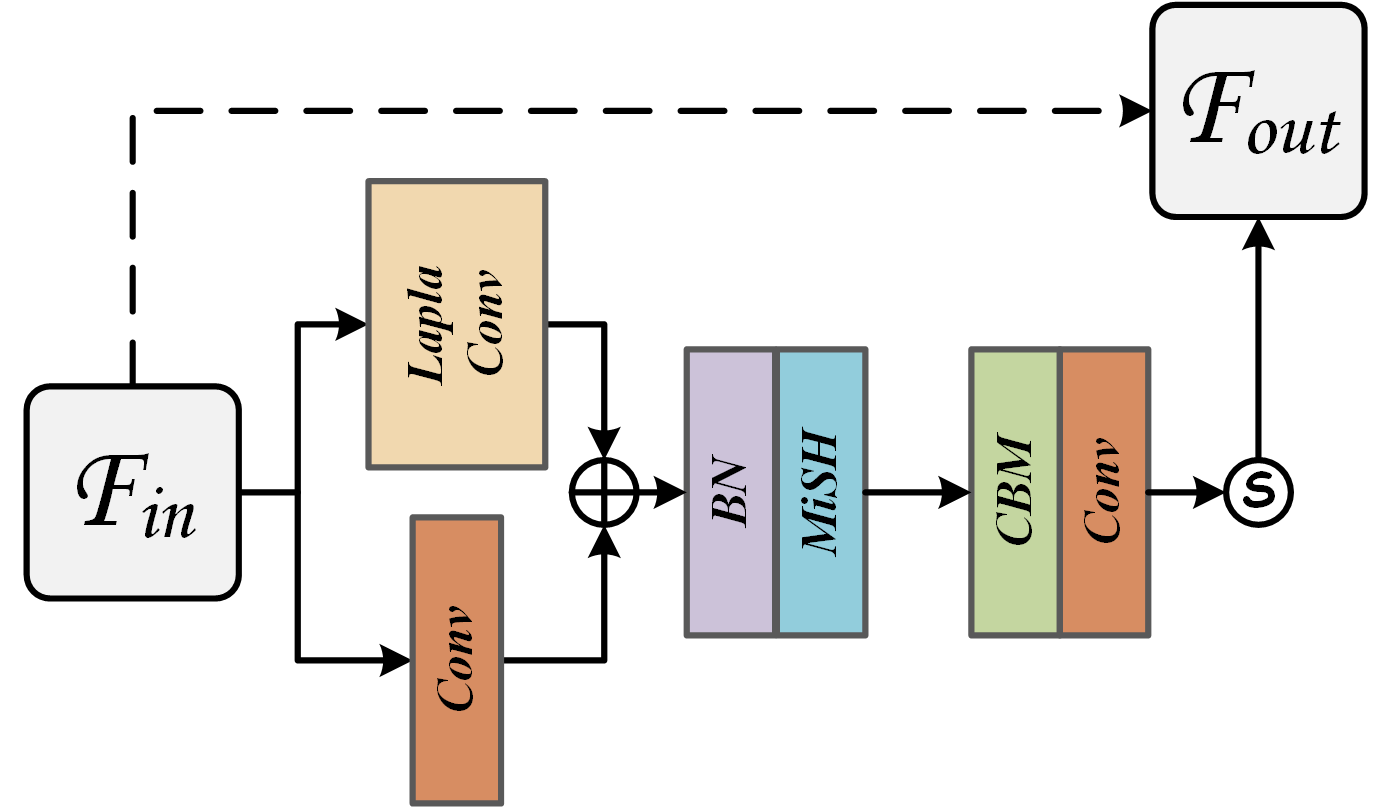}
\caption{Structure of Sense block. Featuring LaplaConv with fixed-weight laplace convolution.}
\label{Fig3}
\end{figure}

We propose a novel Channel Combination Inference Block (CCI) based on this strategy, which infers the true pixels of all three channels by using pairwise combinations between channels. The CCI block's structure is shown in Fig. \ref{FRR_detail} (a). Given an input image \( F_{in} \) of size [B, C, H, W], its convolutional combinations have a receptive field that is neither too local nor too global, matching our strategy of `inferring and restoring primary color pixels from the neighboring regions'. The following equations describe the CCI block:
\begin{align}
& F_{combin} = Concat([F_{in}, F_{in}]) \\
& C_{fix} = GCB_{g=3, N_1}(F_{combin}) \\
& B_{fix} = Concat([C_{fix}, F_{in}])
\end{align}
where GCB denotes $N_1$ grouped convolutional combinations, with the number of groups set to 3, corresponding to the number of channels that need to be inferred (as RGB images have 3 channels); $\sigma$ denotes the Mish activation function. The inferred reconstructed pixels \( C_{fix} \) are concatenated with the original pixels to yield the final output tensor \( B_{fix} \). We opt for concatenation instead of addition because the grouped convolutions inhibit the interchange of gradient information; thus, we cannot fully execute the `inference and restoration of primary color pixels from the neighboring regions.' The network does not truly obtain local features that encompass all three RGB channels, which is specifically validated by subsequent ablation studies. We use concatenation to complete the final restoration in the subsequent stage module, with the entire strategy summarized as follows: 
\begin{align}
& B_{fix} = Concat(R'G'B', RGB) \\
& F_{out} = CBM_{N_2}(B_{fix})
\end{align}
where $N_2$ is the number of CBM block in FSM block. Degradation factors such as irregular noise will be removed after the FRR module.

\subsection{Smoothing High-frequency Laplacian}

UIE is a challenging task because of the complex noise and the lack of clear paired underwater datasets. The water body can also distort the image features when the camera or the photographer moves. This makes some pixels deviate from their actual position, creating a diffusion effect, especially in the deep sea where this phenomenon is more severe. This impairs the faithful representation of pixel features and introduces artifacts, but existing methods cannot cope with it. As a result, the enhancement quality and accuracy for handling highly degraded images are limited.
\begin{figure}[!t]
\centering 
\includegraphics[width=0.47\textwidth]{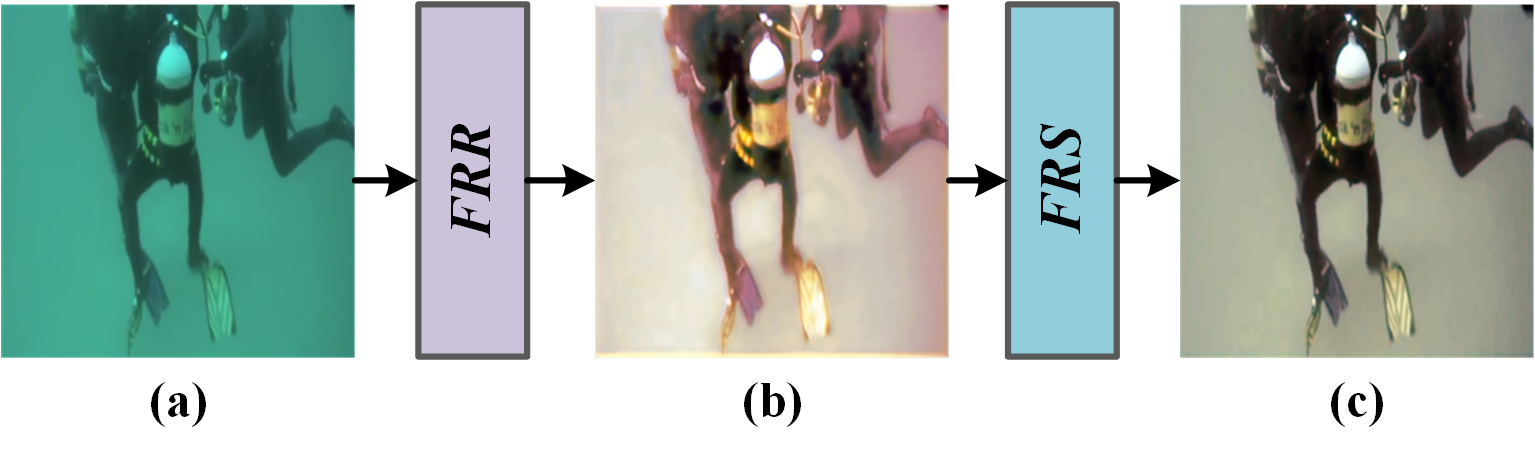}
\caption{Comparative visualization: (a) original image, (b) images processed by FRR module, and (c) images processed by FRS module. (b) and (c) are generated by the proposed DGNet. They could be noticed that the FRS-processed image reduces the artifacts.} 
\label{Fig4}
\end{figure}

To a certain extent, such disturbances in feature distribution can lead to the appearance of artifacts in the enhanced image, as shown in Fig. \ref{Fig4} (b). We propose a new method to smooth the low-frequency regions of the features affected by artifacts, using the Laplacian operator. This method restores the feature distribution to match the real scene, the structure of Sense block (base components of FRS) is shown in Fig. \ref{Fig3}.

The input image \( F_{in} \) is processed through a fixed-weight Laplacian operator convolution kernel, which assigns greater weight to high-frequency features, denoted as:
\begin{align}
& F_{high} = F_{in} \times w \\
& w = \begin{bmatrix}
    0 &  1 & 0  \\
    1 & -4 & 1  \\
    0 &  1 & 0  \\
\end{bmatrix}
\end{align}
where $w$ represents the weight of the Laplacian. 

The obtained high-frequency area is subtracted from a feature map extracted by a 3$\times$3 convolution, using tensor subtraction to generate a low-frequency attention map \( A \). This low-frequency attention map is then sent to a convolutional group to perform feature smoothing. Finally, the low-frequency map, weighted by a Sigmoid function, is added to the original input \( F_{in} \) to produce the final smoothed image \( S \).
\begin{align}
& F_{normal}= CBM(F_{in}) \\
& A = Mish(BN(F_{normal} - F_{High})) \\
& S = F_{in} + Sigmoid(Conv(CBM(A)))
\end{align}

In summary, we analyzed the characteristics of water body oscillations and designed the use of a fixed-weight Laplacian operator to operator to flat-separate high-frequency features in the real world from the spectral domain and smooth the remaining portion of the features. This approach allows for a more accurate interpretation of object edges, ensures smoother transitions of image colors, and solve the problem of edge artifacts after deep-sea image enhancement, as shown in Fig. \ref{Fig4}.

\subsection{Auto-updating Pseudo-labeling Strategy}
Acquiring paired underwater datasets is challenging, and synthetic underwater images lack realism. To overcome the misleading guidance of datasets, we developed a data augmentation method that combines pseudo-labels with paired data. This method applies CLAHE to generate pseudo ground truth (GT) for each predicted image. During the gradient backpropagation process, the pseudo GT produced by the CLAHE-enhanced (CLAHE method helps to improve contrast and balance lighting, which can provide deeper features to the network.) network accentuates features such as light balance, contrast, and more. Simultaneously, it introduces a dynamically guided gradient for image generation, facilitating the network's escape from saddle points in the gradient space and mitigating the negative impact of low-quality GT.

Relying only on pseudo-labeled constraint networks might cause incorrect gradients. Hence, our pseudo-labeling strategy is more like adding a direction marker to the existing gradient space, rather than the self-supervised method of using only pseudo-labels. In particular, we apply CLAHE operations to the network prediction image \( F_{pred} \) to generate pseudo-labels under the current gradient. As a result, the pseudo-label is updated after each gradient descent, and the dynamic gradient it generates evolves. Together with the reference gradient, it forms a continuously updated complete gradient space. This aids the network in building momentum away from saddle points, all the while offering a broader solution space. CLAHE creates pseudo-labels as follows:
\begin{align}
& I_{s} = Split_{t}(I) \label{eq:clahe1} \\
& I_{s}^{e} = HE(I_{s}) \label{eq:clahe2} \\
& E = Concat(I_{s}^{e}) \label{eq:clahe3} \\
& P_{gt} = Clip(E)
\label{eq:clahe4}
\end{align}
where we define the pseudo-labels $P_{gt}$ as the result of applying histogram equalization to the enhanced images $E$, $t$ is set to 8, and $Clip$ restricts the range of the pixel values.

Instead of using the original or the real image, the pseudo-label is based on the predicted image. This approach makes pseudo-label gradient changes as the model weights update, which prevents the network from getting stuck in local optima. Moreover, since CLAHE usually improves the white balance of the image, and white balance is a key feature of the ground image, our pseudo-label also guides the prediction of white balance in multiple training rounds. This enhances the correct gradients that align with the desired white balance while diminishing the impact of incorrect ones. Our experiments show that this label generation method not only speeds up network training but also increases accuracy (detail in supplementary material).

\subsection{Loss Function}
We propose two gradient spaces for guiding the training of the network and improving the accuracy of the network: 

\noindent \textbf{$\bullet$ Reference gradient:} $L_{1}$ loss function is primarily used to assess the prediction error loss between each pixel of the predicted image and the reference image, and it is a part of the reference gradient. $L_{\text{SSIM}}$ loss primarily assesses the similarity \cite{R24SSIM} between the predicted image and the reference ideal image, providing gradients of luminance, contrast, and structure for the reference gradient.

\noindent \textbf{$\bullet$ Dynamic gradient:} Dynamic Tuning Loss $L_{d}^{\tau}$ is primarily used to dynamically adjust the gradient space of the network. It performs the establishment of pseudo-GT on dynamic predicted images and uses MAE (Mean Absolute Error) to evaluate the pixel-level loss between the predicted images and the pseudo-GT images which spawn by Eqs. \eqref{eq:clahe1}, \eqref{eq:clahe2}, \eqref{eq:clahe3}, and \eqref{eq:clahe4}.

\begin{equation}
L = \alpha \cdot L_{1} + \beta \cdot L_{\text{SSIM}} + \gamma \cdot L_{d}^{\tau}
  \label{eq:13}
\end{equation}
where $\alpha$, $\beta$, and $\gamma$ is the weight of each loss.

\begin{figure*}[htb]
\centering 
\includegraphics[width=\textwidth]{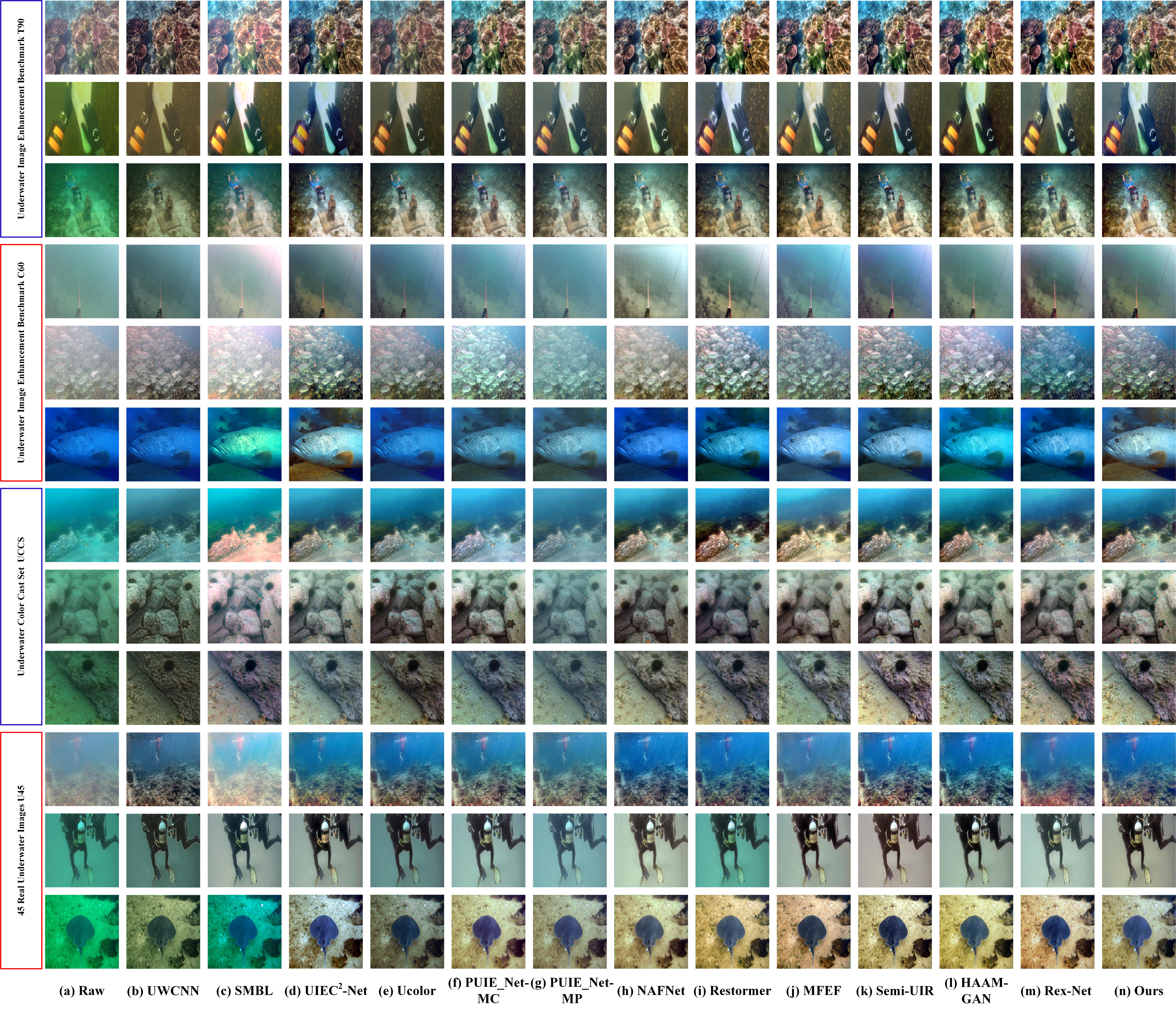}
\caption{Visual comparison of different methods on UIEB, U45, and UCCS datasets. The visual comparison of different methods from left to right includes UWCNN\cite{R28UWCNN}, SMBL\cite{R32SMBL}, UIEC\(^2\)-Net\cite{R26UIEC}, Ucolor\cite{R25Ucolor}, PUIE-Net(MC)\cite{R33PUIENet}, PUIE-Net(MP)\cite{R33PUIENet}, NAFNet\cite{R36NAFNet}, Restormer\cite{R35Restormer}, MFEF\cite{R34MFEM}, Semi-UIR\cite{R37SemiUIR}, HAAM-GAN\cite{R39HAAMGAN}, Rex-Net\cite{R38ReX-Net}, and Our method.}
\label{Fig05}
\end{figure*}







To maximize the effectiveness of the $L_{d}^{\tau}$, during the initial phase of training, the weight of $L_{d}^{\tau}$ should be less than that of $L_{1}$, with the reference gradient dominating the training process. As training progresses, the influence of the reference gradient gradually diminishes, and the dynamic gradient increasingly guides the network. When the network becomes stuck in a local optimum, $L_{d}^{\tau}$ will reconstruct the dynamic gradient based on the current gradient predictions to help the network escape the local optimum without incurring excessive resource costs, thereby quickly converging to a global optimum. Additionally, the dynamic direction of $L_{d}^{\tau}$ always points towards white balance, ensuring that the predicted image's lighting and high-level semantic features on the RGB channels are in line with real-world scenes.

\begin{table*}[htb]
\centering
\resizebox{\linewidth}{!}{
\begin{tabular}{l*{13}{c}}
\toprule
\raisebox{-2ex}[0pt][0pt]{Methods} & \multicolumn{5}{c}{UIEB V90} & \multicolumn{2}{c}{UIEB C60} & \multicolumn{2}{c}{UCCS} & \multicolumn{2}{c}{U45} & \multicolumn{2}{c}{Z700} \\
\cmidrule(lr){2-6} \cmidrule(lr){7-8} \cmidrule(lr){9-10} \cmidrule(lr){11-12} \cmidrule(l){13-14}
 & PSNR$\uparrow$ & RMSE$\downarrow$ & SSIM$\uparrow$ & UCIQE$\uparrow$ & UIQM$\uparrow$ & UCIQE$\uparrow$ & UIQM$\uparrow$ & UCIQE$\uparrow$ & UIQM$\uparrow$ & UCIQE$\uparrow$ & UIQM$\uparrow$ & UCIQE$\uparrow$ & UIQM$\uparrow$ \\
\midrule
SMBL(TB'20)\cite{R32SMBL} & 16.681 & 0.158 & 0.801 & 0.604 & 2.552 & \colorbox{blue!20}{\textbf{0.584}} & 2.006 & 0.549 & 2.552 & 0.602 & 2.416 & 0.561 & 3.084 \\
UWCNN(PR'20)\cite{R28UWCNN} & 17.949 & 0.134 & 0.847 & 0.546 & 3.033 & 0.520 & 2.546 & 0.498 & 3.025 & 0.546 & 3.079 & 0.508 & 2.974 \\
UIEC\(^2\)-Net(SPIC'21)\cite{R26UIEC} & 22.958 & 0.078 & 0.907 & 0.600 & 3.013 & \colorbox{blue!20}{\textbf{0.584}} & 2.611 & 0.544 & 2.992 & 0.608 & 3.145 & 0.577 & 2.996 \\
Ucolor(TIP'21)\cite{R25Ucolor} & 21.093 & 0.096 & 0.872 & 0.580 & 3.048 & 0.553 & 2.482 & 0.550 & 3.019 & 0.573 & 3.159 & 0.550 & 3.059 \\
PUIE-Net (MC)(ECCV'22)\cite{R33PUIENet} & 21.382 & 0.093 & 0.882 & 0.582 & 3.049 & 0.558 & 2.521 & 0.536 & 3.003 & 0.578 & 3.199 & 0.562 & 3.075 \\
PUIE-Net (MP)(ECCV'22)\cite{R33PUIENet} & 18.257 & 0.126 & 0.778 & 0.569 & 3.064 & 0.513 & 2.398 & 0.489 & 2.758 & 0.563 & 3.186 & 0.548 & 3.083 \\
NAFNet(ECCV'22)\cite{R36NAFNet} & 22.691 & 0.080 & 0.870 & 0.592 & 3.044 & 0.559 & \colorbox{blue!20}{\textbf{2.751}} & 0.544 & 2.913 & 0.594 & 3.087 & 0.571 & 3.011 \\
Restormer(CVPR'22)\cite{R35Restormer} & 23.701 & 0.073 & 0.907 & 0.599 & 3.015 & 0.570 & 2.688 & 0.546 & 2.980 & 0.600 & 3.097 & 0.576 & \colorbox{blue!20}{\textbf{3.073}} \\
MFEF(EAAI'23)\cite{R34MFEM} & 23.589 & 0.073 & 0.918 & 0.600 & 3.062 & 0.566 & 2.652 & 0.556 & 2.977 & 0.606 & 3.157 & 0.584 & 2.981 \\
Semi-UIR(CVPR'23)\cite{R37SemiUIR} & 24.309 & 0.073 & 0.901 & 0.605 & 3.032 & 0.583 & 2.663 & 0.557 & \colorbox{blue!20}{\textbf{3.079}} & 0.606 & 3.185 & 0.591 & 2.900 \\
ReX-Net(ESWA'23)\cite{R38ReX-Net} & 23.734 & 0.073 & 0.910 & 0.598 & 2.980 & 0.570 & 2.654 & 0.540 & 2.897 & 0.597 & 3.128 & 0.573 & 3.058 \\
HAAM-GAN(EAAI'23)\cite{R39HAAMGAN} & 22.824 & 0.080 & 0.882 & 0.602 & 3.169 & 0.570 & \colorbox{red!20}{\textbf{2.876}} & 0.556 & 3.029 & 0.606 & 3.172 & 0.578 & 2.998 \\
\midrule
Our-S & \colorbox{blue!20}{\textbf{25.067}} & \colorbox{blue!20}{\textbf{0.071}} & \colorbox{blue!20}{\textbf{0.925}} & \colorbox{red!20}{\textbf{0.611}} & \colorbox{red!20}{\textbf{3.100}} & \colorbox{red!20}{\textbf{0.592}} & 2.731 & \colorbox{blue!20}{\textbf{0.565}} & 3.016 & \colorbox{blue!20}{\textbf{0.607}} & \colorbox{red!20}{\textbf{3.219}} & \colorbox{blue!20}{\textbf{0.595}} & \colorbox{red!20}{\textbf{3.089}} \\
Our-L & \colorbox{red!20}{\textbf{25.623}} & \colorbox{red!20}{\textbf{0.069}} & \colorbox{red!20}{\textbf{0.929}} & \colorbox{blue!20}{\textbf{0.610}} & \colorbox{blue!20}{\textbf{3.081}} & \colorbox{red!20}{\textbf{0.592}} & 2.736 & \colorbox{red!20}{\textbf{0.569}} & 3.011 & \colorbox{red!20}{\textbf{0.610}} & \colorbox{blue!20}{\textbf{3.212}} & \colorbox{red!20}{\textbf{0.593}} & 3.061 \\
\bottomrule
\end{tabular}}
\caption{Quantitative method comparisons across UIEB, UCCS, U45, and Z700 datasets. Top performances highlighted: best in Red, second in blue.}
\label{T1}
\end{table*}

\section{Experiments}
In this section, detailed information about the experimental setup and analysis of the comparison with state-of-the-art methods is provided. More details on experiments, such as color constancy verification experiments based on the Grey World Theory \cite{R54color}, are provided in the Supplementary Materials.

\subsection{Implementation Details}
In our implementation, the proposed DGNet is divided into two models: Our-s (a small model with $N_1=3$ and $N_2=3$) and Our-l (a large model with $N_1=6$ and $N_2=6$) designed for different computational devices and task scenarios, as shown in Fig. \ref{FRR_detail}. The experiments were conducted on a system running Ubuntu 20.04 with an Intel Xeon Gold 6330 CPU and an RTX 3090 24GB GPU. The software package was built using Python 3.10 and PyTorch 2.1.0 with CUDA 11.8 support.

The Exponential Moving Average (EMA) strategy was employed, and the input size was gradually increased from 256$\times$256 to 400$\times$400 during training to enhance the model's generalization capability. The Adamw optimizer was chosen, and the learning rate was set to 1e-3 with a batch size of 5 under global seed 0.

\subsection{Datasets and Evaluation Metrics}

We used four publicly available datasets as benchmarks in our experiments: UIEB~\cite{R18WaterNet}, U45~\cite{R20}, UCCS~\cite{R19}, and Z700. The UIEB dataset consisted of 890 real underwater images with synthetic references and 60 unpaired real images (C60). The U45 and Z700 datasets contained 45 and 700 realistic underwater scenes, respectively. The UCCS dataset comprised real images from three scenes: 100 blue-shifted, 100 green-shifted, and 100 blue-green-shifted. We divided the paired data in the UIEB dataset into a training set of 800 images (T800) and a validation set of 90 images (V90). T800 served as the training set, V90 as the validation set, and the other datasets were used as test sets for all the methods.

To comprehensively evaluate the performance of the proposed DGNet, we adopted three reference metrics: Mean Squared Error (MSE), Peak Signal-to-Noise Ratio (PSNR)~\cite{R23PSNR}, and Structural Similarity (SSIM)~\cite{R24SSIM}, to measure the differences between the enhanced images and the reference images. Additionally, we employed two widely used no-reference metrics: Underwater Image Quality Measure (UIQM)~\cite{R22UIQM} and Underwater Color Image Quality Evaluation Metric (UCIQE)~\cite{R21UCIQE}, to quantify color smoothness, clarity, and contrast. To further assess the practical performance of the model, we measured the frame rate (The FPS results are displayed in the Supplementary Material.) for inferring HD images on an RTX 3090, which indicates the inference speed of the model.

\subsection{Comparision with SOTA Methods}
\subsubsection{Qualitative Comparisons}
Visual comparisons of our method (Our-L) and 12 other SOTA methods are presented in Fig. \ref{Fig05}. We selected representative scenes from four datasets that cover a wide range of real underwater scenarios. Our method effectively corrects color casts and exhibits higher saturation in the V90 dataset. Our method outperforms other methods in terms of preserving details in the C60 dataset. The second row demonstrates the consistent performance of our method in haze removal for objects at different distances. The third row shows that our method, along with UIEC\(^2\)-Net, achieves better restoration results for the given scene. Our method proves to be effective in various color-shifted scenes in the UCCS dataset. In the first scene of U45, only our method, Semi-UIR, HAAM-GAN, and Rex-Net reasonably restore the red and green colors of the bottom vegetation. However, HAAM-GAN exhibits grid patterns, while Rex-Net shows overcompensation in red. In the second scene, our method successfully restores object edges and color shifts. In the third scene, our method accurately restores the original colors of the beach, while the other methods exhibit unrealistic color rendering. 
\vspace{-1em}
\subsubsection{Quantitative Comparisons}
As displayed in Tab. \ref{T1}, our method achieves a quantitative comparison with 16 SOTA methods. Both the Our-S and Our-L models obtain higher values across various metrics, significantly surpassing other SOTA methods. On the UIEB C90 dataset, compared to Semi-UIR (CVPR'23), our Our-L models demonstrate improvements of 5.4\% in PSNR and 3.1\% in SSIM, respectively. Additionally, as observed from Fig. \ref{Fig1}, our method exhibits lower parameter size and computational complexity, making it suitable for practical applications.
\begin{table}
\centering
\resizebox{0.9\linewidth}{!}{
\begin{tabular}{c|c|c c c c}
\hline
\multicolumn{2}{c|}{\textbf{Operation}} & \textbf{PSNR} & \textbf{SSIM} & \textbf{UIQM} & \textbf{UCIQE} \\ \hline
\multicolumn{2}{c|}{Instead ALL} & 15.892 & 0.732 & 1.780 & 0.595 \\ \hline
\multicolumn{1}{c|}{\raisebox{-5ex}[0pt][0pt]{FRR}} 
& w/o CCI & 25.231 & \colorbox{blue!20}{\textbf{0.925}} & 3.073 & 0.608 \\
& w/o FSM & 23.164 & 0.916 & 3.055 & 0.601 \\
& w/o Sigmoid & 24.869 & 0.922 & 3.045 & 0.604 \\
& Remove & 24.916 & 0.922 & \colorbox{blue!20}{\textbf{3.077}} & 0.610 \\
& Instead & 16.468 & 0.715 & 1.538 & 0.620 \\
\hline
\multicolumn{1}{c|}{\raisebox{-4ex}[0pt][0pt]{FRS}} 
& w/o Lapla & \colorbox{blue!20}{\textbf{25.256}} & 0.924 & 3.065 & 0.610 \\
& w/o SenB & 25.111 & 0.923 & 3.051 & 0.610 \\
& Remove & 25.040 & 0.920 & 3.051 & \colorbox{red!20}{\textbf{0.611}} \\
& Instead & 16.111 & 0.721 & 1.597 & 0.610 \\
\hline
\multicolumn{2}{c|}{ALL} & \colorbox{red!20}{\textbf{25.623}} & \colorbox{red!20}{\textbf{0.929}} & \colorbox{red!20}{\textbf{3.081}} & \colorbox{blue!20}{\textbf{0.610}} \\
\hline
\end{tabular}
}
\caption{Ablation for the FRR and FRS modules and all their sub-modules. The \textbf{Remove} indicates deletion from the full model, and the \textbf{Instead} denotes replacement with a CBM group of equivalent parameters and receptive fields.
}
\label{tab:FRRS}
\end{table}

\subsection{Ablation Study}
\textbf{\begin{figure}[htb]
\centering
\includegraphics[width=3in]{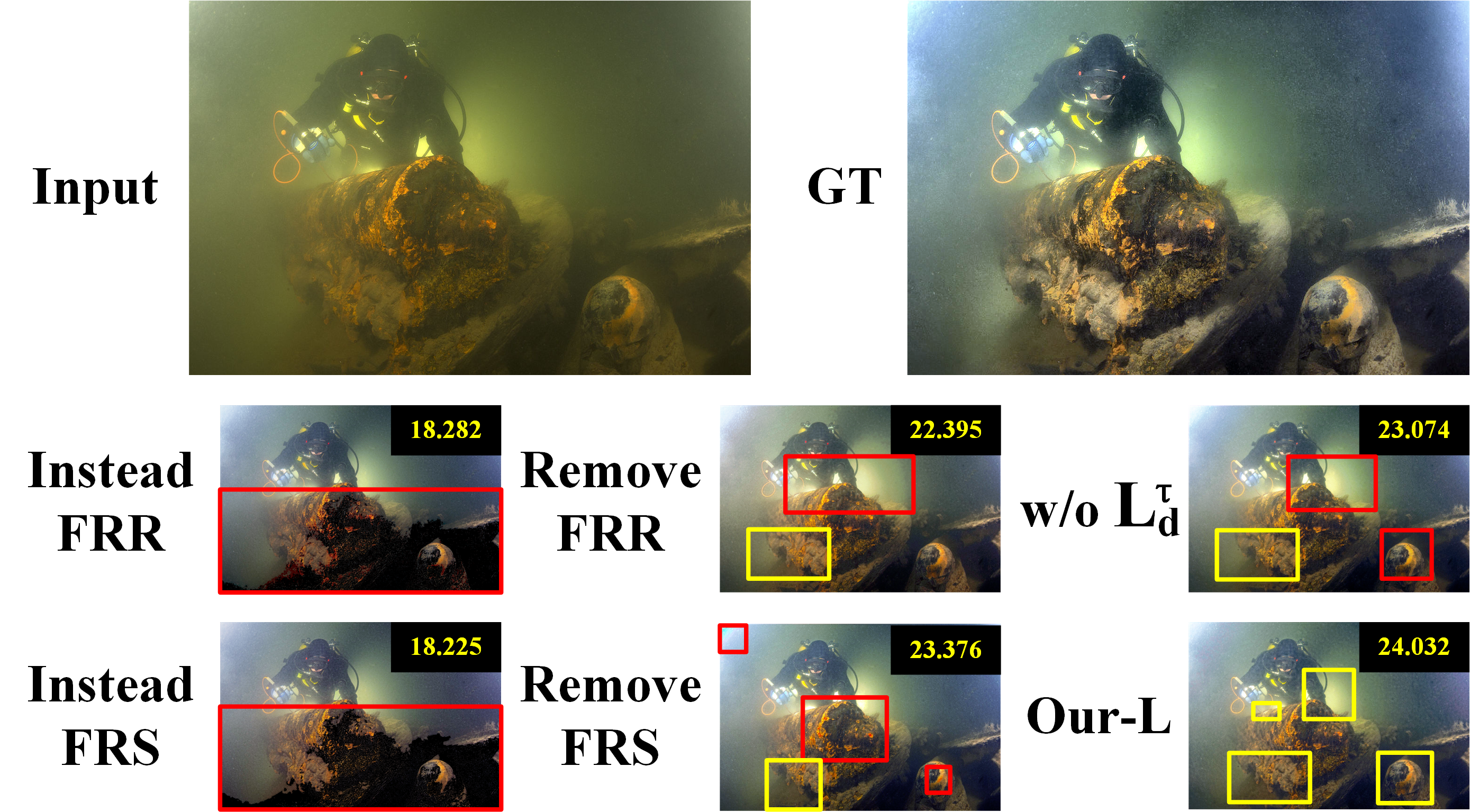}
\caption{Visual comparison of ablation study on UIEB V90. The \textcolor[RGB]{220, 0, 0}{red box} shows the error enhancements, the \textcolor[RGB]{220, 220, 0}{yellow box} shows the excellent enhancements and the PSNR score is in the upper right corner of the image.}
\label{Fig:Ablation}
\end{figure}
}

\subsubsection{Effectiveness of FRR}
Our FRR module is designed to reduce noise and balance the colors in images. As depicted in Figure \ref{Fig4} (a) and (b), it effectively removes color casts but introduces motion blur. The results of the ablation study in Table \ref{tab:FRRS} reveal that FRR not only improves feature repair and reconstruction but also enhances performance, increasing PSNR by 0.7 and SSIM by 0.007.

\subsubsection{Effectiveness of FRS}
The FRS module, which employs fixed-weight Laplacian convolution, excels at extracting low-frequency features and reducing artifacts. Table \ref{tab:FRRS} demonstrates that FRS significantly boosts PSNR and SSIM by 11.5 and 0.2, respectively, surpassing other convolution groups. The `Remove' experiment further validates the role of FRS in enhancing model performance. Additionally, experiments involving the Laplacian convolution and smoothing module confirm the effectiveness of our approach. A comparison between Figure \ref{Fig4} (b) and \ref{Fig:Ablation} (c) clearly illustrates the FRS module's ability to eliminate blurring and artifacts, particularly at object edges.

\begin{table}[htb]
\centering
\resizebox{0.8\linewidth}{!}{
\begin{tabular}{c|cccc}
\hline
\multicolumn{1}{c|}{\textbf{Loss}} & \textbf{PSNR} & \textbf{SSIM} & \textbf{UIQM} & \textbf{UCIQE} \\ \hline
w/o $L_{1}$ & 25.133 & \colorbox{red!20}{\textbf{0.931}} & 3.053 & 0.608 \\ \hline
w/o $L_{ssim}$ & 25.201 & 0.912 & \colorbox{red!20}{\textbf{3.125}} & \colorbox{red!20}{\textbf{0.611}} \\ \hline
w/o $L_{d}^{\tau}$ & 25.404 & 0.927 & 3.064 & 0.610 \\ \hline
\multicolumn{1}{c|}{ALL} & \colorbox{red!20}{\textbf{25.623}} & \colorbox{blue!20}{\textbf{0.929}} & \colorbox{blue!20}{\textbf{3.081}} & \colorbox{blue!20}{\textbf{0.610}}\\
\hline
\end{tabular}
}
\caption{Evaluating combinations of Loss functions.}
\label{tab:dym}
\end{table}

\subsubsection{Effectiveness of $L_{d}^{\tau}$}
Our $L_{d}^{\tau}$ is designed to preserve advanced features such as contrast and light balance, expedite network training, and assist the network in avoiding local optima. As demonstrated in Tables \ref{tab:dym} and \ref{tab:lossw}, an increase in the weight of $L_{d}^{\tau}$ correlates with a rise in UIQM values, indicating its effective role in guiding contrast and light balance. Moreover, the PSNR value is 0.24 dB higher when using $L_{d}^{\tau}$ compared to not using it, illustrating its ability to guide the network more effectively towards the global optimum. As shown in Figure \ref{Fig:Ablation}, $L_{d}^{\tau}$ also allows the images recovered by our model to outperform the light and contrast balance of the ground truth to some extent.

\begin{table}[htb]
\centering
\resizebox{0.7\linewidth}{!}{
\begin{tabular}{c|cccc}
\hline
\multicolumn{1}{c|}{\textbf{Weight}} & \textbf{PSNR} & \textbf{SSIM} & \textbf{UIQM} & \textbf{UCIQE} \\ \hline
0.00 & 25.133 & \colorbox{red!20}{0.931} & 3.053 & \colorbox{blue!20}{\textbf{0.608}} \\ \hline
0.35 & \colorbox{red!20}{\textbf{25.623}} & \colorbox{blue!20}{\textbf{0.929}} & 3.081 & \colorbox{red!20}{\textbf{0.610}} \\ \hline
0.70 & \colorbox{blue!20}{\textbf{25.522}} & 0.928 & 3.093 & 0.607 \\ \hline
1.00 & 24.742 & 0.925 & \colorbox{blue!20}{\textbf{3.112}} & 0.605 \\ \hline
1.50 & 17.030 & 0.736 & \colorbox{red!20}{\textbf{3.377}} & 0.590 \\
\hline
\end{tabular}
}
\caption{Assessing the impact of $L_{d}^{\tau}$ weight variations.}
\label{tab:lossw}
\end{table}
\section{Conclusion}
In this study, we propose a lightweight and efficient enhancement network for underwater image enhancement, improves visual quality, particularly in noise reduction, effectively boosting the model's generalizability and adaptability. Our proposed dynamic gradient guidance strategy effectively controls the restoration effect, providing a novel perspective on the interpretability of UIE tasks. Comparative experiments demonstrate that our method surpasses the Semi-UIR method (CVPR’23): achieving a higher PSNR of 25.1 (a 3.3\% improvement), 88.3\% fewer parameters, and a 61.7\% increase in computational efficiency, making it more suitable for underwater exploration. 
\medskip
{
    \small
    \bibliographystyle{ieeenat_fullname}
    \bibliography{arxiv}
}


\end{document}